\documentclass[sigconf]{acmart}

\usepackage{setspace}
\usepackage{multirow}
\usepackage{makecell}

\AtBeginDocument{%
  \providecommand\BibTeX{{%
    \normalfont B\kern-0.5em{\scshape i\kern-0.25em b}\kern-0.8em\TeX}}}

\copyrightyear{2022}
\acmYear{2022}
\setcopyright{acmcopyright}
\acmConference[MM '22] {Proceedings of the 30th ACM International Conference on Multimedia }{October 10--14, 2022}{Lisboa, Portugal.}
\acmBooktitle{Proceedings of the 30th ACM International Conference on Multimedia (MM '22), Oct. 10--14, 2022, Lisboa, Portugal}
\acmPrice{15.00}
\acmISBN{978-1-4503-9203-7/22/10}
\acmDOI{10.1145/3503161.3548148}

\begin{document}

\title{Mutual Adaptive Reasoning for Monocular 3D Multi-Person Pose Estimation}



\author{Juze Zhang$^{1,2,3}$,
Jingya Wang$^{1,4 *}$,
Ye Shi$^{1,4 *}$,
Fei Gao$^1$,
Lan Xu $^{1,4}$,
Jingyi Yu $^{1,4}$}
\affiliation{
$^1$ School of Information Science and Technology, ShanghaiTech University, Shanghai 201210, China\\
$^2$ Shanghai Advanced Research Institute, Chinese Academy of Sciences, Shanghai 201203, China\\
$^3$ University of Chinese Academy of Sciences, Beijing 100049, China\\
$^4$ Shanghai Engineering Research Center of Intelligent Vision and Imaging
	\city{}
	\country{}
}
\email{
{zhangjz,wangjingya,shiye,gaofei,xulan1,yujingyi}@shanghaitech.edu.cn
}


\renewcommand{\shortauthors}{Juze Zhang et al.}

\begin{abstract}
Inter-person occlusion and depth ambiguity make estimating the 3D poses of monocular multiple persons as camera-centric coordinates a challenging problem. Typical top-down frameworks suffer from high computational redundancy with an additional detection stage. By contrast, the bottom-up methods enjoy low computational costs as they are less affected by the number of humans.  However, most existing bottom-up methods treat camera-centric 3D human pose estimation as two unrelated subtasks: 2.5D pose estimation and camera-centric depth estimation. In this paper, we propose a unified model that leverages the mutual benefits of both these subtasks. Within the framework, a robust structured 2.5D pose estimation is designed to recognize inter-person occlusion based on depth relationships. Additionally, we develop an end-to-end geometry-aware depth reasoning method that exploits the mutual benefits of both 2.5D pose and camera-centric root depths. This method first uses 2.5D pose and geometry information to infer camera-centric root depths in a forward pass, and then exploits the root depths to further improve representation learning of 2.5D pose estimation in a backward pass. Further, we designed an adaptive fusion scheme that leverages both visual perception and body geometry to alleviate inherent depth ambiguity issues. Extensive experiments demonstrate the superiority of our proposed model over a wide range of bottom-up methods. Our accuracy is even competitive with top-down counterparts. Notably, our model runs much faster than existing bottom-up and top-down methods. 
\end{abstract}




\thanks{*Corresponding author}

\maketitle

\begin{figure}
\centering
	\includegraphics[width=1.0\columnwidth]{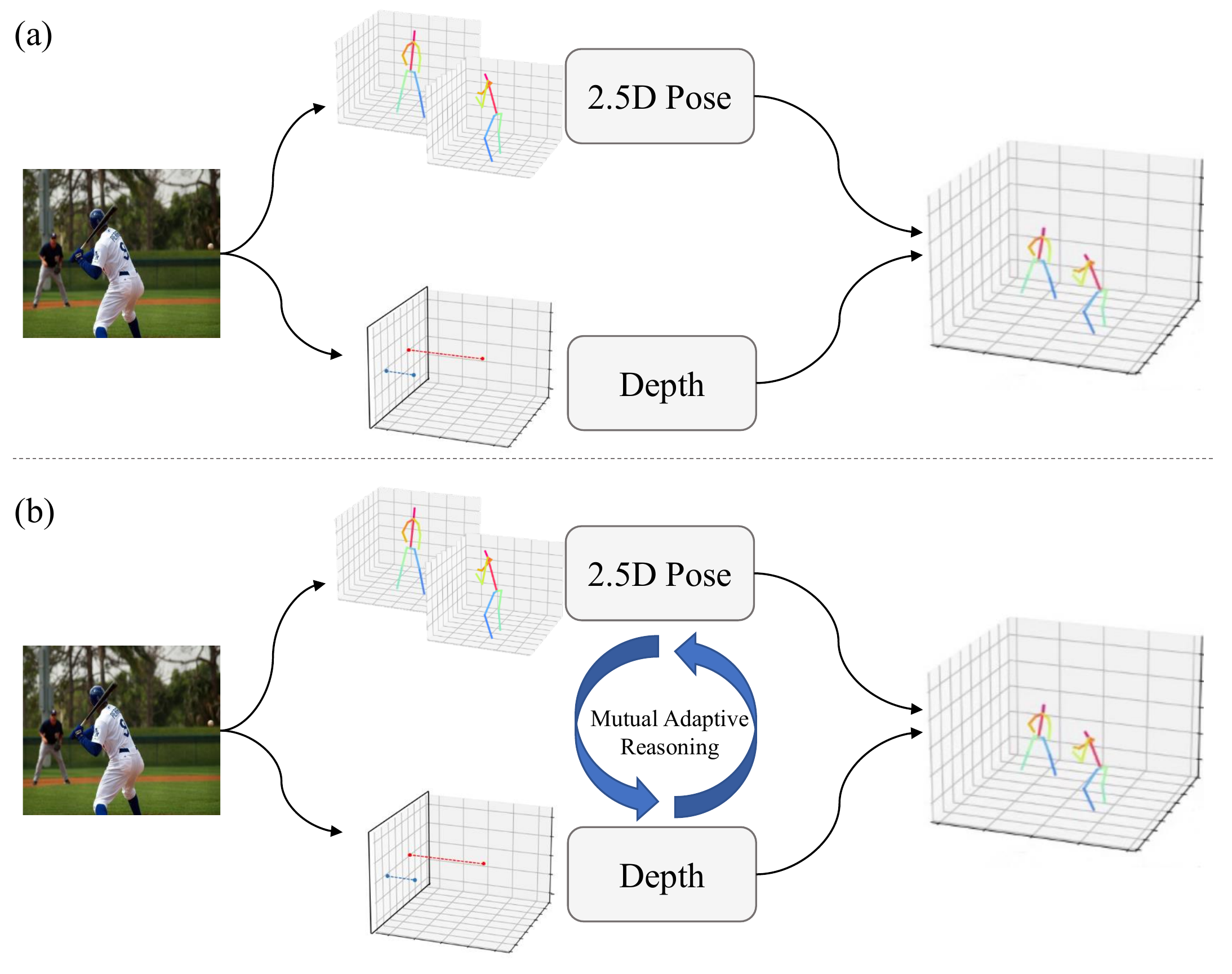}
\caption{(a) Existing methods treat camera-centric 3D-MPE as two unrelated subtasks: 2.5D pose representation and absolute depth estimation. (b) Our method can bridge the gap between 2.5D pose representation and depth estimation and achieve mutual benefit from them.}
\label{figure1}
\vspace{-2em} 
\end{figure}

\section{Introduction}
In recent years, 3D pose estimation has attracted a great deal of interest from researchers due to its widespread application in a range of fields, including video analysis, camera surveillance, human-computer interaction, virtual/augmented reality, etc. Great successes have been achieved in monocular 3D human pose estimation, especially when dealing with a single person in an image\cite{sun2018integral, zeng2021learning, kocabas2019self, iqbal2020weakly,cheng2019occlusion,wandt2019repnet}. However, when it comes to real-world scenarios, inferring poses can be much harder, especially when multiple persons are present in the scene. This is because body joints are often occluded by other objects or people. For this reason, monocular 3D multi-person pose estimation (3D-MPE) is still a challenging problem, yet, robustness to such occlusions is critical to real-world applications. 

Camera-centric 3D-MPE tasks aim to recover each pose as coordinates in the camera-centric coordinate system. This requires estimating the absolute depth of each person in 3D. Generally, existing methods for 3D-MPE can be divided into two categories: top-down and bottom-up. Typical top-down approaches use the single-person method with a 2D person detector to handle multi-person scenes \cite{moon2019camera}. This involves a human detector and pose estimation in two stages. Since each person in the image is treated individually, these methods ignore out-of-patch contexts. They also suffer from high computational redundancy. By contrast, bottom-up 3D-MPE \cite{zhen2020smap} does not need human detectors and can perceive global image contexts in a single shot. Thus, multi-person interactions in pose estimation tasks do not present a problem. 

\begin{figure*}
\centering
	\includegraphics[width=2.0\columnwidth]{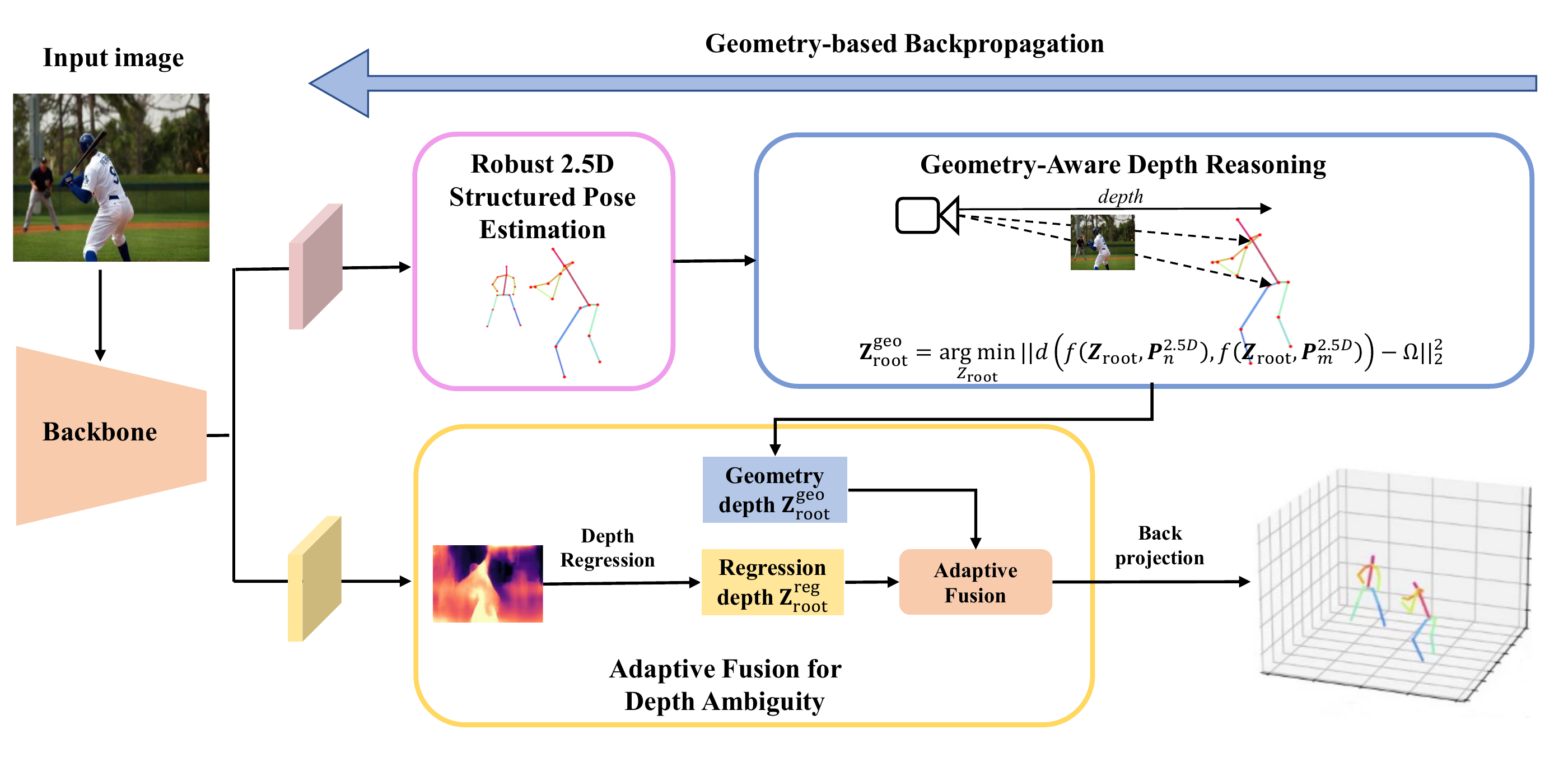}
\caption{Overview of the proposed Mutual Adaptive Reasoning (MAR) framework.}
\label{Overview}
\vspace{-1em} 
\end{figure*}

Most existing bottom-up methods treat camera-centric 3D-MPE as two unrelated subtasks: 2.5D pose representation and depth estimation as shown in Figure \ref{figure1}a. We argue that a robust 2.5D pose representation, especially in crowded multi-person scenes, requires the global depth cues over the whole image to be aggregated for disambiguation. Further, the geometric information associated with 2.5D poses can lead to a closed-form solution for individual depths. This observation motivated us to design adaptive reasoning between pose and depth estimation that mutually benefit each other. To this end, we developed a novel unified model that combines the two subtasks as shown in Figure \ref{figure1}b.
To improve the representation learning of 2.5D pose estimation, we developed an end-to-end method using the information of geometry depths. To disambiguate the depths, we designed a depth fusion scheme that leverages both direct visual perception and geometry to deal with inherent depth ambiguity issues in monocular human depth estimation. Thus, the model can bridge the gap between 2.5D pose representation and depth estimation. Moreover, the framework includes a process for robust structured pose refinement to handle occlusive or out-of-image joints. This means the model is capable of end-to-end learning without needing an additional network in the post-processing stage to complete any missing joints. The whole pipeline is illustrated in figure \ref{Overview}.
To summarize, our main contributions are:
\begin{itemize}
	\item A unified bottom-up model that leverages the mutual benefits of 2.5D pose and depth estimation. Exploiting each strength yields greater robustness for disambiguation. 
	\item An end-to-end geometry aware depth reasoning method that first uses 2.5D pose and geometry information to infer camera-centric root depths in a forward pass, and then exploits the root depths to further improve representation learning of 2.5D pose estimation in a backward pass. 
	\item An adaptive depth fusion that leverages both direct visual perception and geometry to deal with inherent depth ambiguity issues in monocular human depth estimation.
\end{itemize}

\section{Related Work}
\textbf{Multi-Person 2D Pose Estimation.} As mentioned, existing methods for multi-person 2D pose estimation can mainly be divided into top-down and bottom-up approaches. Typical top-down frameworks deal with human detection and pose estimation in two stages \cite{chen2018cascaded,fang2017rmpe}. Despite their strength at handling scale variation, top-down methods also suffer from high computational redundancy when attempting to detect additional persons. Moreover, their performance tends to degrade severely with heavy occlusion since they have no awareness of out-of-patch contexts. Bottom-up approaches \cite{cao2017realtime,cao2021openpose,newell2017associative, hidalgo2019single,nie2019single,papandreou2018personlab,braso2021center,geng2021bottom, zhou2019objects, li2019rethinking, jin2020differentiable} localize all key points in the image first and then group them to an individual. Examples include OpenPose \cite{cao2017realtime}, a representative work, which was the first to present the part-affinity field approach that links key points likely to lie in the same person. Zhou et al. \cite{zhou2019objects} were the first to propose a single-stage offset-based method for 2D human pose estimation. However, the above method only focuses on the 2D level and is careless about 3D information from the image.

\noindent\textbf{Top-Down Multi-Person 3D Pose Estimation.} 
Early works focused on human-centric tasks by using a top-down manner without estimating individual depth \cite{rogez2017lcr,rogez2019lcr}. Only a few works tackle the problem of camera-centric multi-person 3D pose estimation from a monocular RGB image or video. Moon et al. \cite{moon2019camera} was the first to suggest locating a person’s root absolute depth by learning a correction factor of the area of 2D bounding box, while Lin et al. \cite{lin2020hdnet} proposed a pose-aware human depth estimation network to address the problem of root joint localization for multi-person 3D pose estimation in the camera-space. Similarly, Veges et al. \cite{veges2019absolute} proposed to use two separate networks, pose estimator and depth estimator, to recover camera-centric multi-person 3D pose. To handle inherent depth ambiguity, Wang et al. \cite{wang2020hmor} proposed a novel hierarchical multi-person ordinal relation. Another type of work utilizes temporal information to recover 3D poses from a given video\cite{cheng2021graph,cheng2021monocular}. By applying a top-down scheme, the above method either directly regresses the absolute 3D depth from a cropped image, or it computes it based on a prior of the body size, ignoring global image contexts. Moreover, they suffer from high computational redundancy with an additional detection stage. By contrast, our methods can utilize global image contexts and enjoy low computational costs as they are less affected by the number of humans.




\noindent\textbf{Bottom-Up Multi-Person 3D Pose Estimation.}
 A few bottom-up methods have been proposed for 3D-MPE\cite{mehta2018single, fabbri2020compressed,mehta2020xnect,zhen2020smap}. Mehta et al. \cite{mehta2018single} first proposed a bottom-up method with an occlusion-robust pose-map (ORPM) to represent the occlusive joints. Such methods can alleviate the occlusion problem to some degree, but they only focus on person-centric and do not infer multiple persons as camera-centric coordinates. While Fabbri et al. \cite{fabbri2020compressed} used an encoder-decoder network to compress a heatmap and then decompress it back to the original resolution in the inference time for fast HD image processing. They show superior results when tackling crowded scenes but careless about the occlusion problem. Another method called XNect \cite{mehta2020xnect}, a framework that encodes a 3D location map at the spatial location of each visible joint, which also focuses on a person-centric problem. Zhen et al. \cite{zhen2020smap} proposed an end-to-end network with the depth-aware part association and bone-length constraints that benefits the 2.5D pose estimation branch of the scheme. However, these models heavily depend on post-processing techniques or refinement. Above bottom-up methods treat camera-centric 3D human pose estimation as two unrelated subtasks: 2.5D pose estimation and camera-centric depth estimation. By contrast, our model focus on exploring the mutual benefits 2.5D pose estimation and camera-centric depths in an end-to-end manner.

\section{Method}
Our task is to estimate the 3D poses for multiple people in a monocular RGB image as camera-centric coordinates considering the challenge of inter-person occlusion and depth ambiguity. In this paper, we propose a Mutual Adaptive Reasoning (MAR) method for this task. MAR is a unified bottom-up model that leverages the mutual benefits of 2.5D pose and depth estimation. Within the model, a robust structured 2.5D pose estimation is first designed to recognize inter-person occlusion based on depth relationships. Additionally, we develop an end-to-end geometry aware depth reasoning method that exploits the mutual benefits of both 2.5D pose and camera-centric root depths. This method first uses 2.5D pose and geometry information to infer camera-centric root depths in a forward pass, and then exploits the root depths to further improve representation learning of 2.5D pose estimation in a backward pass. Further, we designed a depth fusion scheme that leverages both visual perception and body geometry to alleviate inherent depth ambiguity issues. An overview of the proposed Mutual Adaptive Reasoning (MAR) method is depicted in Figure \ref{Overview}. More details on the structured pose estimation module are provided in Section \ref{sec:Robust}, followed by the direct depth regression module in Section \ref{sec:Direct}, the end-to-end geometry depth aware reasoning module in Section \ref{sec:Differentiable} and the adaptive fusion for depth ambiguity in Section\ref{sec:fusion}.


\subsection{Robust 2.5D Structured Pose Estimation}
\label{sec:Robust}
For monocular 3D multi-person pose estimation (3D-MPE), the goal is to recover the absolute camera-centric coordinates of the keypoint of multiple people $\{{\bf{P}}^{3D}_k \}^J_{k=1}$, where ${\bf{P}}^{3D}_k = [X_k,Y_k,Z_k]^T $ and $J$ represent the number of joints. A 3D pose can be decomposed into representations of 2.5D pose $\{{\bf{P}}^{2.5D}_k \}^J_{k=1} = \{\left[u_k,v_k,Z_k^r\right]^T\}^J_{k=1} $ and the absolute depth of the root joint $Z_{\text{root}}$. Here, the coordinates $u_k$ and $v_k$ are the image pixel coordinates of the $k$-th keypoint, $Z_k^r$ is its relative depth to the root keypoint and $Z_{\text{root}}$ is the absolute camera-centric depth of each person. Given a 2.5D pose $\{{\bf{P}}^{2.5D}_k \}^J_{k=1}$, several elements can be derived from back-projection. These are: the intrinsic camera parameters $K\in \mathbb{R}^{3\times 3}$; the absolute human depth $Z_{\text{root}}$, and the final absolute 3D pose $\{{\bf{P}}^{3D}_k \}^J_{k=1}$. Let $f: ( {\bf{P}}^{2.5D}_k ,\bf{Z}_{\text{root}})\rightarrow {\bf{P}}^{3D}_k $ be a back-projection function. Thus,
\begin{equation}
f : {\bf{P}}^{3D}_k = \left( \bf{Z}_{\text{root}} + Z_k^r\right)K^{-1}[u_k,v_k,1]^T, \label{back-project}
\end{equation}
where
$$
K =\left[ \begin{array}{ccc}
f_x & 0 & c_x\\
0 & f_y & c_y\\
0 & 0 & 1
\end{array} \right ],
$$ $f_x,f_y$ is focal lengths divided by the per-pixel distance factors (pixel) of x-axis and y-axis and $c_x,c_y$ is principal point of x-axis and y-axis.

\begin{figure}
\centering
	\includegraphics[width=1.0\columnwidth]{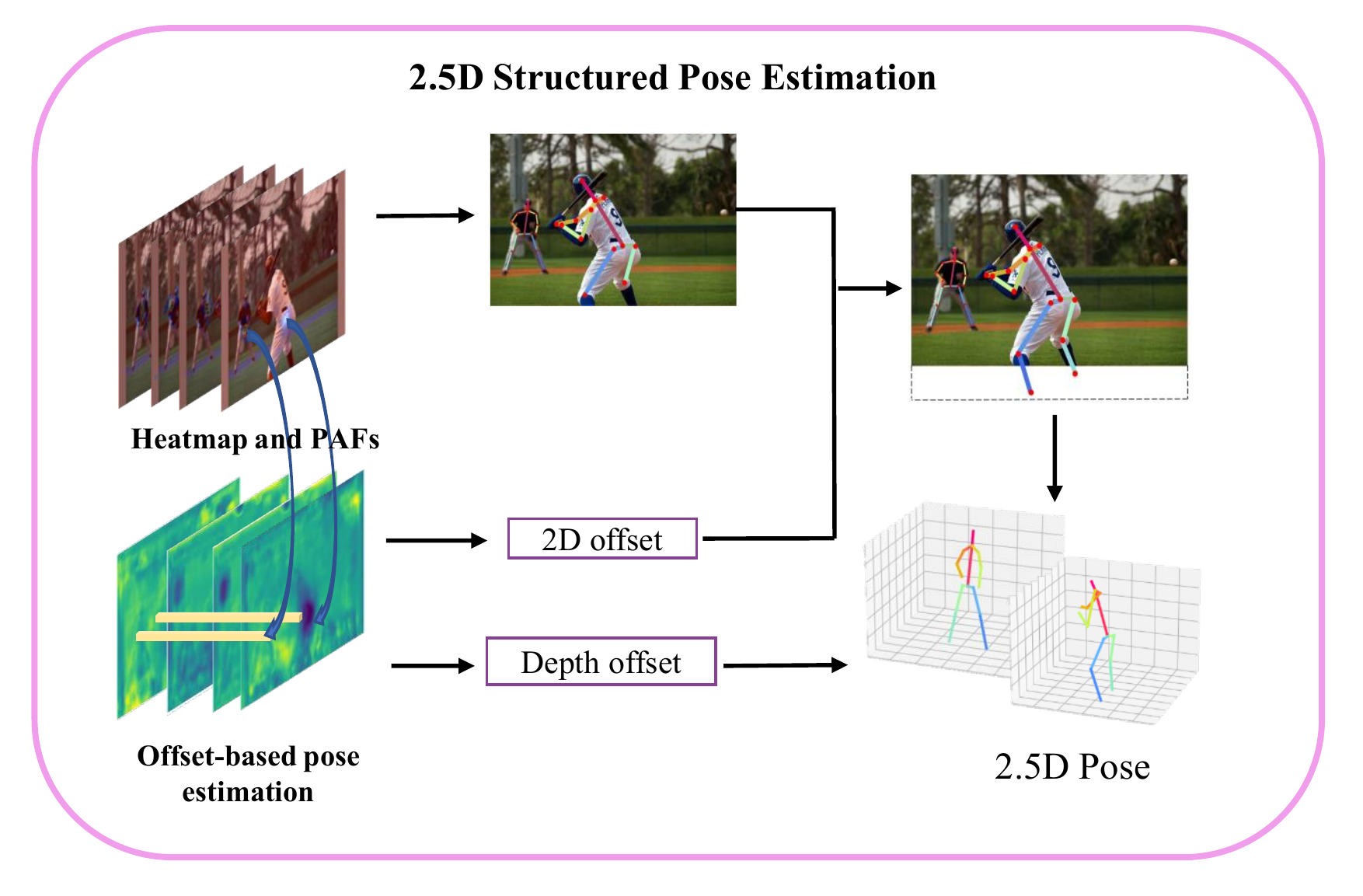}
\caption{Robust 2.5D Structured Pose Estimation}
\label{pose}
\vspace{-2em}
\end{figure}

In real application scenarios, the body joints may be occluded, making it hard to infer occluded joints from heatmaps. Thus, we designed the offset scheme for pose estimation, which handles issues of occlusion as well as completely out-of-frame joints as shown in Fig.\ref{pose}. Appearance-based heatmap representations have a more precise spatial positioning accuracy, but they usually fail to infer occluded joints. By contrast, 2.5D joint offset scheme estimates its position from the offset between root joints, which is more robust to occlusive joints. Here, we propose a robust structured pose estimation that simultaneously estimates heatmaps and 2.5D joint offsets.


A heatmap of keypoints encodes the center of a 2D human body in the image, where the activation value at each position indicates the degree to which the keypoint lies in position. The position of each joint is represented as a gaussian distribution in the heatmap, and we use PAFs to link the identity-agnostic joints from the keypoint heatmaps after non-maximum suppression following \cite{ding2020learning}. 
Then, we can get the  2D joints location $(u_k^{hm}, v^{hm}_k)^T$ at image pixel coordinate from heatmap scheme.

Moreover, we introduce 2.5D joint offset scheme that aims to estimate $(J-1)$ human joint offsets with respect to a root joint. The offset maps are encoded at the spatial location of each root joint predicted by the keypoint heatmap. Formally, the 2.5D offset-based pose estimation is formulated as follows:
\begin{equation}
(u_k^{disp},v_k^{disp}, Z_k^r) = (u_\text{root},v_\text{root},0) + (\Delta_{u_k}, \Delta_{v_k}, \Delta_{ Z_k}),  
\end{equation}
where $(u_\text{root},v_\text{root},0)$ represents the spatial location of the root joint as image coordinates and $(\Delta_{u_k}, \Delta_{v_k}, \Delta_{ Z_k})$ represents the offset of the $k$-th body joint position with respect to the root joint. Note that the relative depth of the root joint with respect to itself must be zero. Based on above observation, we use heatmap and part affinity fields to estimate the major keypoints. Occlusive or out-of-image joints can be inferred using the offset scheme. The final structured 2.5D pose ${\bf{P}}^{2.5D}_k$ is then inferred as follows: 
\begin{equation}
{\bf{P}}^{2.5D}_k = \begin{cases}
(u_k^{hm}, v^{hm}_k, Z_k^r)^T & \text{if}\ c_k \geq \text{threshold},\\
(u_k^{disp},v_k^{disp}, Z_k^r)^T &
\text{else},
\end{cases}
\label{2.5Dinfer}
\end{equation}
where $c_k$ is the local maximum value of heatmaps for joint $k$. It is worth noting that Eq.\ref{2.5Dinfer} was only used in inference stage. The 2.5D pose estimation is trained in an end-to-end manner with the depth estimation. The total training loss of the pose estimation includes heatmap loss, PAF loss and keypoints offset loss.

\subsection{Depth Regression} \label{sec:Direct}
Generally, the human depth estimated from a single camera view is inherently ambiguous. This ambiguity will make it hard to learn an accurate regression model. To handle this issue, we exploit the capability of probability prediction to capture the tolerance of depth ambiguity by predicting a depth distribution instead of an absolute depth, which has also been used in the field of object detection \cite{he2019bounding}. Here, we assume that the depth distribution of each person is independent and follows a Laplace distribution $L({z}^{\text{reg}}_{\text{root}},\  \lambda)$.  The probability density function of a Laplace random variable $Z \sim L({Z}^{\text{reg}}_{\text{root}},\  \lambda)$ is:
\begin{equation}
p_{\text{reg}}(Z) = \frac{1}{2\lambda} e^{-\frac{| Z - {Z}^{\text{reg}}_{\text{root}} |}{\lambda}}
\end{equation}
where $\mu$ and $\lambda$ are parameters of the Laplace distribution.
The ground-truth depth can also be formulated as a Laplace distribution. Since it is deterministic, it can be further represented by a Dirac delta function:
\begin{equation}
p_{D}(Z) = \delta(Z - \hat{Z}_{\text{root}}),
\end{equation}
where $\hat{Z}_{\text{root}}$ represents the ground truth of depth. The distance between the distribution of direct depth and the ground truth is measured by the Kullback Leibler(KL) Divergence. 
\begin{align}
{L}_{\text{reg}} & = D_{KL}(p_{D}(Z)\|p_{\text{reg}}(Z) ) \notag \\ 
& \propto \frac{|{Z}^{\text{reg}}_{\text{root}} - \hat{Z}_{\text{root}}|_1}{\sigma_1} + \log(\sigma_1), 
\end{align}
where $\sigma_1 = \sqrt{2}\lambda$ is the standard deviation of the Laplace distribution to indicate the depth uncertainty. If the model lacks confidence in its prediction, it will output a larger $\sigma$ so that $L_{\text{reg}}$ can be reduced. The term $\log(\sigma)$ avoids trivial solutions and encourages the model to be optimistic about accurate predictions. Detailed derivation is provided in Supplementary.

\subsection{Geometry-Aware Depth Reasoning} \label{sec:Differentiable}
The inherent ambiguity makes it hard to learn an accurate regression model. Given the difficulty to regress depth directly, we develop a geometry depth reasoning method that exploits the mutual benefits of both 2.5D pose and camera-centric root depths. This method first uses 2.5D pose and geometry information to infer camera-centric root depths in a forward pass, and then exploits the root depths to improve representation learning of 2.5D pose estimation in a backward pass.


\subsubsection{Geometry-based Forward Pass}
Once the 2.5D pose ${\bf{P}}^{2.5D}$ has been determined by Section 3.1, the 2D image coordinates should be back-projected to the camera-centred coordinate space using the estimated depth value to get the final coordinates of ${\bf{P}}^{3D}$. 
Given a 2.5D pose ${\bf{P}}^{2.5D}$, the intrinsic camera parameters $K$, and a torso length $\Omega$, the geometry reasoning depth can be inferred as follows:
\begin{equation}
{Z}^{\text{geo}}_{\text{root}} = \mathop{\arg\min}\limits_{{Z}_{\text{root}}} \| d(f({Z_{\text{root}},{\bf{P}}^{2.5D}_{\text{n}}),f({Z}_{\text{root}},{\bf{P}}^{2.5D}_{\text{m}})) - \Omega\|_2^2}
\label{argmin}
\end{equation}
where the function $d(\cdot,\cdot)$ measures the distance between the \textit{root joint} and \textit{neck joint} in the camera-centric space. Eq.\ref{argmin} leads to a closed-form solution as follows:
\begin{equation}
{Z}^{\text{geo}}_{\text{root}} = \frac{-b + \sqrt{b^2 - 4ac}}{2a},\label{closed}
\end{equation}
where
$$
a=(f_x^{-1}\Delta_{u_m})^2 + (f_y^{-1}\Delta_{v_m})^2,
$$
$$
b= 2 Z_m^r [f_x^{-2} \Delta_{u_m} (u_m -c_x) + f_y^{-2} \Delta_{v_m} (v_m -c_y) ],
$$
$$
c= (Z_m^r)^2 [f_x^{-2} (u_m -c_x)^2 + f_y^{-2} (v_m -c_y)^2 + 1]^2 - \Omega^2.
$$ Detailed derivation is provided in Supplementary. To optimize the final geometry aware depth distribution, we apply the uncertainty regression loss same as \ref{sec:Direct}:

\begin{equation}
{L}_{\text{geo}} =  \frac{|{Z}^{\text{geo}}_{\text{root}} - \hat{Z}_{\text{root}}|}{\sigma_2 }+ \log(\sigma_2). 
\end{equation} 

\subsubsection{Geometry-based Backward Pass}
To further improve the representation learning of 2.5D pose estimation, we develop an end-to-end differentiable method using the geometry depth information. The geometry depth, which is deduced by using 2.5D pose, can be used to further update the feature extractor in a backward process.  
Since the geometry-based root depth  ${\bf{Z}}^{\text{geo}}_{\text{root}}$ can expressed in a closed-form solution with respect to the 2.5D pose ${\bf{P}}^{2.5D}_k$ as shown in Eq.\ref{closed}, we can directly compute the backward derivative as follows: 
%
\begin{eqnarray}
&&\frac{\partial {L}_{\text{geo}}}{\partial {\bf{P}}^{2.5D}_k} = \frac{\partial {L}_{\text{geo}}}{\partial {Z}^{\text{geo}}_{\text{root}}}  \frac{\partial {Z}^{\text{geo}}_{\text{root}}}{\partial {\bf{P}}^{2.5D}_k} \nonumber\\
&&= ({Z^{\text{geo}}_{\text{root}} - \hat{Z}_{\text{root}}) \cdot (\frac{\partial {Z}^{\text{geo}}_{\text{root}}}{\partial a} \frac{\partial a}{\partial {\bf{P}}^{2.5D}_k}} \nonumber\\
&&+ \frac{\partial {Z}^{\text{geo}}_{\text{root}}}{\partial b} \frac{\partial b}{\partial {\bf{P}}^{2.5D}_k}  +\frac{\partial {Z}^{\text{geo}}_{\text{root}}}{\partial c} \frac{\partial c}{\partial {\bf{P}}^{2.5D}_k}). 
\end{eqnarray}

\subsection{Adaptive Fusion for Depth Disambiguation}
\label{sec:fusion}
Existing approaches either use a neural network to regress the 3D depth of a person from the cropped image based on the body size, or they directly regress the depth based on visual perception of the object scale.  However, both methods focus on the local and global cues independently; they do not consider any synergies between them. Moreover, direct depth regression will result in satisfactory depth estimation, but that result may be inaccurate in some occluded or crowded scenes. A key insight into solving the depth ambiguity issue is, therefore, to design a mutual adaptive reasoning depth directly from the visual perception that also includes a geometry-based reasoning depth. Doing so should also bridge the gap between 2.5D pose representation and depth estimation. Therefore, the loss function to train the whole network is as follows:
\begin{equation}
{L} = {L}_{\text{geo}} + \omega {L}_{\text{reg}},
\end{equation}
where $\omega$ is a hyperparameter to trade-off between the geometry loss and regression loss.



At this point, we have two depths estimated from direct depth regression and geometry reasoning. The uncertainty of the direct regressed depth and the geometry aware depth is used to fuse the two depths, expressed as follows:
\begin{equation}
{Z}^{\text{fusion}}_{\text{root}} = (\frac{{Z}_{\text{root}}^{\text{reg}}}{\sigma_1} +\frac{ {Z}^{\text{geo}}_{\text{root}}}{\sigma_2} )/ (\frac{1}{\sigma_1} +\frac{ 1}{\sigma_2} ),
\end{equation}
where $\sigma_1$ is the direct regress uncertainty and $\sigma_2$ is the geometry reasoning uncertainty.

\section{Experiments}

\begin{table*}[htbp]
\centering
\begin{spacing}{1.15}
\begin{tabular}{c|c|llll|ll}
\hline
\multicolumn{1}{l}{}       &        & \multicolumn{4}{c}{All people}         \vline             & \multicolumn{2}{c}{Matched} \\
\hline
\multicolumn{1}{l}{}    \makecell[{}{p{1.5cm}}]{Scheme}  \vline & Method & $\text{PCK}_\text{rel}$ & $\text{PCK}_\text{abs}$ & $\text{PCK}_\text{root}$  & PCOD  & $\text{PCK}_\text{rel}$ & $\text{PCK}_\text{abs}$\\
\hline
\multirow{7}{*}{Top-down}& Rogez et al. \cite{rogez2017lcr}   &\makecell{53.8} & \makecell{-}   & \makecell{-} & \makecell{-} & \makecell{62.4} & \makecell{-}   \\
                         & Rogez et al. \cite{rogez2019lcr}   &\makecell{70.6} & \makecell{-}   & \makecell{-} & \makecell{-} & \makecell{74.0} & \makecell{-}   \\
                         & Dabral et al. \cite{dabral2019multi}&\makecell{71.3} & \makecell{-}   & \makecell{-} & \makecell{-} & \makecell{74.2} & \makecell{-}    \\
                         & Moon et al. \cite{moon2019camera} &\makecell{81.8} &\makecell{31.5} & \makecell{-} & \makecell{92.6} & \makecell{82.5} & \makecell{31.8}  \\
                         & Lin et al. \cite{lin2020hdnet}   & \makecell{-}   & \makecell{-}   & \makecell{- }&\makecell{ -} &\makecell{ 83.7} &\makecell{ 35.2}  \\
                          & Wang et al. \cite{wang2020hmor}   & \makecell{-}   &\makecell{ -}   &\makecell{ -} &\makecell{ -} & \makecell{82.0} &\makecell{ \bf{43.8}}  \\
                        & Guo et al. \cite{guo2021pi}   & \makecell{\bf{82.5}}   &\makecell{ -}   &\makecell{ -} & \makecell{-} &\makecell{ \bf{83.9}} &\makecell{ 35.2 } \\
\hline
\multirow{6}{*}{Bottom-up} & Mehta et al. \cite{mehta2018single}  & \makecell{65.0} &\makecell{   - } & \makecell{ - }  & \makecell{  -}  & \makecell{ 69.8} & \makecell{ -}     \\
                           & Mehta et al. \cite{mehta2020xnect}   & \makecell{70.4} & \makecell{  -}  & \makecell{ - }  &\makecell{  - }  &  \makecell{75.8} &\makecell{  - }       \\
                           & Benzine et al. \cite{benzine2020pandanet}& \makecell{72.0} & \makecell{-} & \makecell{ -}   & \makecell{ - }  &\makecell{  -} &\makecell{  -}  \\
                           &Benzine et al. \cite{benzine2021single}& \makecell{67.5} & \makecell{19.8} & \makecell{ -}   &\makecell{  -}   &\makecell{  72.7} &\makecell{  20.9}  \\
                           &Zhen et al. \cite{zhen2020smap}    & \makecell{73.5 }&\makecell{ 35.4} &\makecell{ 45.5} & \makecell{97.0 }& \makecell{ 80.5} & \makecell{ 38.7}     \\
                           & \bf{Ours}            & \makecell{\bf{76.2}} & \makecell{\bf{36.3}} &\makecell{ \bf{48.9}} &\makecell{   \bf{97.1}}  & \makecell{ \bf{81.5}} &\makecell{  \bf{39.5}}     \\
\hline
\end{tabular}
\caption{Results on the MuPoTS-3D dataset. All numbers are average values over 20 activities. “-” means the results that are not available. Best in \bf{bold}.}
\label{tab:MuPoTS}
\end{spacing}
\vspace{-3em} 
\end{table*}

\subsection{Datasets and Evaluation Metrics}

\noindent\textbf{MuCo-3DHP and MuPoTS Datasets.} MuCo-3DHP and MuPoTS-3D, two datasets proposed by \cite{mehta2018single}, were used to assess the framework’s ability to estimate multi-person 3D poses. MuCo-3DHP, the training set we used is a large-scale synthesized dataset, which was generated from single person 3D pose estimation dataset MPI-INF-3DHP \cite{mehta2017monocular} by randomly compositing the persons. We use the same set of MuCo-3DHP synthesized images from \cite{moon2019camera} for a fair comparison. 400K frames of MuCo-3DHP are used for training, among which half are background augmented. We used MuPoTS-3D datasets as our test set which is a real-world outdoor scenes dataset captured by a marker-less motion capture system. Besides, an additional 2D human keypoint dataset COCO is used to train together with the MuCo-3DHP dataset following Mehta et al. \cite{moon2019camera}. Accordingly, we set the loss value of depth becomes zero when the COCO dataset was imported.


\noindent\textbf{Human3.6M Dataset.} The Human3.6M dataset \cite{ionescu2013human3} is currently the largest publicly available dataset for human 3D pose estimation. Two experimental protocols are widely used for training and testing. \emph{Protocol} 1 uses S1, S5, S6, S7, S8, S9 in training and S11 in testing, while  \emph{Protocol} 2 uses S1, S5, S6, S7, S8 in training and S9, S11 in testing. Same as the configuration of MuCo-3DHP dataset, additional 2D human keypoint dataset COCO is used to train together with the Human3.6M dataset. Accordingly, we set the loss value of depth becomes zero when COCO dataset was imported. Following previous work \cite{moon2019camera}, we use \emph{Protocol} 2 and sample every 5th and 64th frames in videos for training and testing respectively.


\noindent\textbf{COCO Dataset.} The COCO dataset \cite{lin2014microsoft} contains over 250, 000 person instances labeled and 200,000 images with 17 keypoints annotations. COCO dataset is divided into three sets named train, val and test-dev, containing 57k, 5k and 20k images respectively. In this paper, we used train2017 as an additional 2D human keypoint dataset. Accordingly, we set the loss value of depth becomes zero followed previous work \cite{moon2019camera} when COCO dataset was imported.



\noindent\textbf{Evaluation Metrics.} Although our task is 3D-MPE at camera-centric coordinates, we also perform person-centric 3D-MPE evaluation metrics.  We use Percentage of Correct 3D Keypoints($\text{PCK}$) to evaluate the performance of 3D-MPE on MuPoTS-3D, which calculate the percentage of correct joints if it lies within 15cm from the ground truth joint location. Following \cite{moon2019camera}, We report the relative 3D$\text{PCK}_\text{rel}$ that with root alignment to evaluate the person-centric 3D-MPE, and absolute 3D$\text{PCK}_\text{abs}$ that without root alignment to evaluate the camera-centric 3D-MPE. To compare the human depth location ability, we evaluate $\text{PCK}_\text{root}$ that only measures the accuracy of root joints and percentage of correct ordinal depth (PCOD) that measures the accuracy of ordinal depth following \cite{zhen2020smap}. 

\subsection{Implementation Details}
We use HRNet-w32 \cite{sun2019deep,wang2020deep} pre-trained on the ImageNet dataset as our backbone network and Adam as our optimizer with a $5\times 10^{-4}$ learning rate and a $10^{-6}$ weight decay. All input images were padded to the same size of 512 × 832. Every prediction head attached to the backbone consists of one $3 \times 3 \times 256$ conv layer, BatchNorm, ReLU and another $1 \times 1 \times c_0$ conv layer, where $c_0$ is the output size. Following \cite{chen2018cascaded}, we implemented multi-scale supervision from the four scale output of HRNet-w32 at the heatmap and part affinity field head. The model was trained for 50k iterations with a batch size of 32 on four RTX 3090 GPUs; $50\%$ of data in each mini-batch was from COCO dataset \cite{lin2014microsoft}. The data augmentation process during the training included rotation, horizontal flips, and color jittering.



\subsection{Quantitative Evaluation on MuPoTS-3D}
To evaluate the performance of 3D-MPE in complex scenarios, we perform experiments on MuPoTS-3D and compare it to current state-of-the-art methods, as shown in Table \ref{tab:MuPoTS}. After root alignment with the ground-truth poses, our model achieves higher performance and outperforms the SOTA\cite{zhen2020smap} on $\text{PCK}_\text{rel}$ by $1.3\%$. The results show that our person-centric 3D-MPE outperforms the SOTA on $\text{PCK}_\text{rel}$ by $2.7\%$ so that our structured 2.5D Pose can significantly improve the performance of the 2.5D pose. Additionally, our method is superior to the most top-down methods and comparable with the SOTA top-down methods on person-centric 3D-MPE metrics for matched ground truths. Since the existing top-down methods require an additional SOTA human detector, such as Mask R-CNN model \cite{he2017mask}, the metrics on $\text{PCK}_\text{rel}$ show an advantage for recognizing instances, but at the expense of computational complexity. We also perform camera-centric evaluation using $\text{PCK}_\text{abs}$, where we outperform the SOTA by 0.9\%. To evaluate our root joint localization performance, we additionally calculate $\text{PCK}_\text{root}$ to evaluate the 3D localization of people. Notably, we achieve better $\text{PCK}_\text{root}$ than SOTA by a large margin, which shows our method can alleviate inherent depth ambiguity issues. Furthermore, our method can achieve 97.1\% on PCOD. It clearly indicates our method can correctly identify the ordinal depth and we analyze this mainly with the help of geometric information. The evaluation on MuPoTS-3D shows that our method outperforms the state-of-the-art methods in both 3D pose and 2.5D pose performance.

\subsection{Quantitative Evaluation on Human3.6M}
The Human3.6M dataset has been widely used to evaluate 3D single person pose estimation. As our task is focused on camera-centric 3D-MPE to handle inter-person occlusion and depth ambiguity, we do not expect our method to perform significantly better than the SOTA methods. The MRPE of root joint localization results on Human3.6M dataset are shown in Table \ref{tab:Human3.6M}. The top 3 rows are baseline results that take from the figures reported in \cite{lin2020hdnet}. The baseline methods \cite{mehta2017monocular,rogez2017lcr} employed an optimization process to minimize the error of back-projection between 2D and 3D pose. “w/o limb joints” refers to optimization using only head and body trunk joints. “with RANSAC” refers to randomly sampling the set of joints used for optimization with RANSAC. 
As shown in Table 1, our method is slightly superior to the SOTA top-down method \cite{lin2020hdnet}. The main reason is that Human3.6M is an indoor single-person dataset, which is quite clean with almost no crowded scenarios. Notably, \cite{moon2019camera} and \cite{lin2020hdnet} are used Mask-Rcnn \cite{he2017mask} with the help of large detection datasets, which resulting our method got slightly lower accuracy along x- and y-axis. Despite this, our method shows better depth location ability at camera-centric coordinates.

\begin{figure*}
\centering
	\includegraphics[width=1.9\columnwidth]{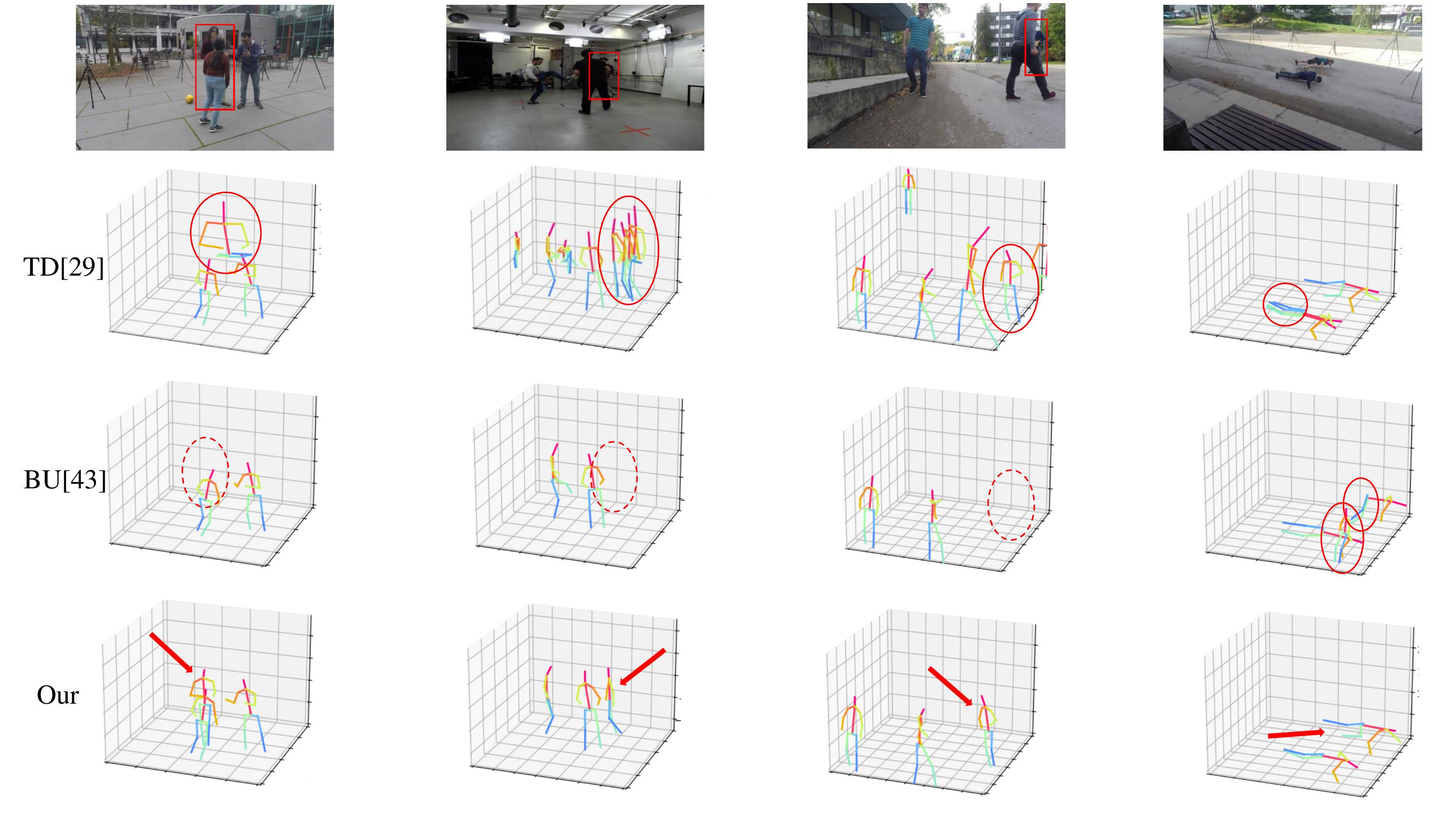}
\caption{Quality results comparisons with SOTA top-down and bottom-up methods.TD stand for top-down method \cite{moon2019camera}, BU stand for bottom-up method \cite{zhen2020smap}. The first row shows the images from four challenging frames from MuPoTS-3D which included extreme occlusion cases and extreme hard poses; the second row shows the results from Posenet\cite{moon2019camera}; the third row shows the result from SMAP \cite{zhen2020smap}; the last row shows the results of our method. The occlusive person is highlighted with red bounding box. Wrong estimations are highlighted with red solid circles. Missing estimations are highlighted with red dotted circles. Correct estimations are highlighted with red arrows.}
\label{result-comparion}
\end{figure*} 

\subsection{Qualitative Comparisons on  MuPoTS-3D and COCO}

\begin{figure*}
\centering
	\includegraphics[width=1.9\columnwidth]{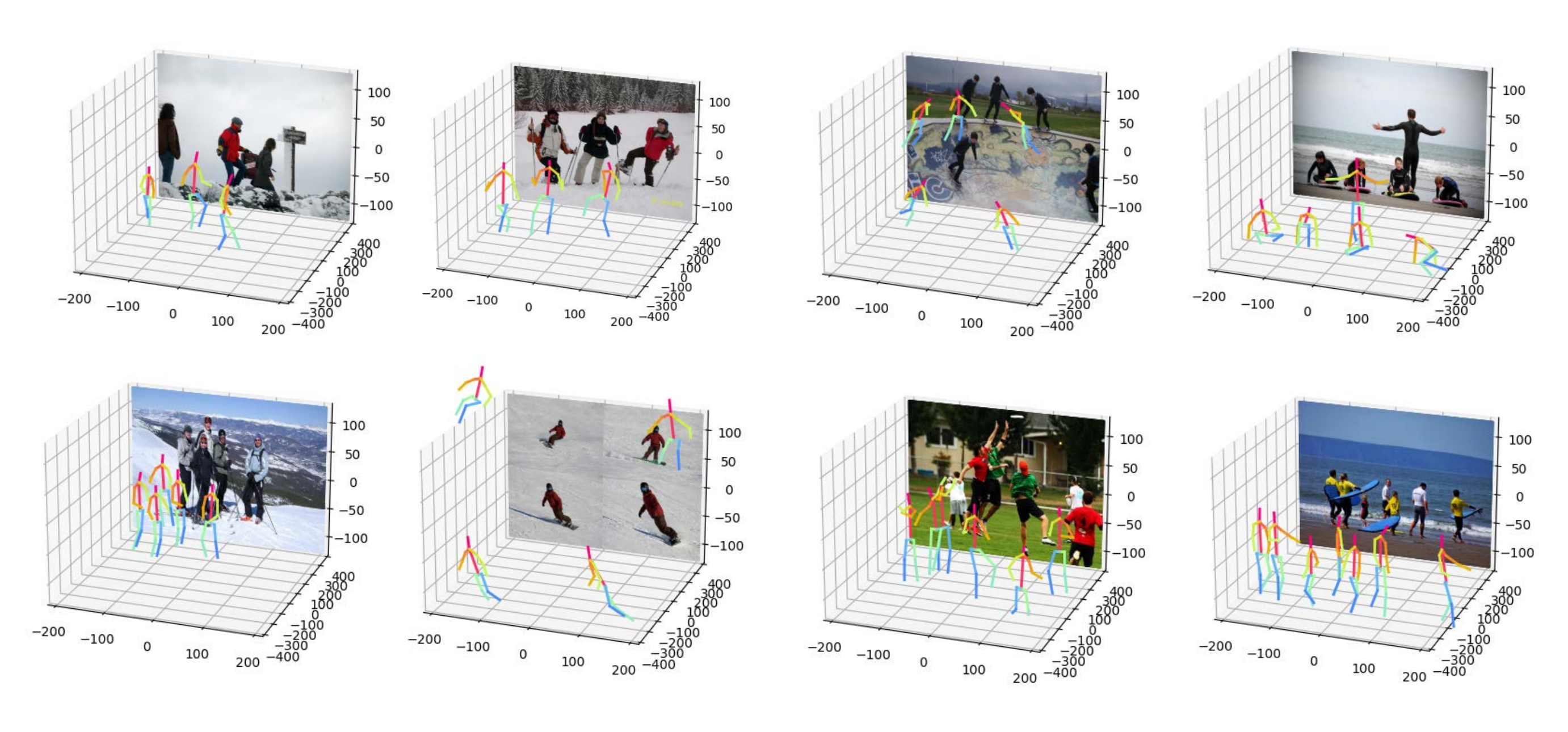}
\caption{The quality results on in-the-wild images from COCO dataset.}
\vspace{-0.5em} 
\label{quality}
\end{figure*}

 Fig. \ref{result-comparion} shows the comparison among a typical top-down method Postnet\cite{moon2019camera}, a SOTA bottom-up method SMAP \cite{zhen2020smap} and our method. We selected four challenging frames from  MuPoTS-3D which included extreme occlusion cases and extreme hard poses. As shown in the second row of Fig. \ref{result-comparion}, we observe that although the top-down method achieves high accuracy with PCK metrics, they suffer from high redundancy for the human detection bounding box and cannot handle well in occlusion case. Specifically, when a person is partially occluded, high redundancy detection bounding box will greatly affect the performance. Besides, SOTA bottom-up method suffers from person scale variation and also cannot handle partial occlusion cases. As shown in the third row of Fig. \ref{result-comparion}, we observe that they cannot detect the partially occluded person behind as well as cannot accurately regress extreme hard poses. By contrast, our method can produce reasonable relative poses and reasonable depth in the above cases as shown in the fourth row of Fig. \ref{result-comparion}. We also provide results of the quality results on in-the-wild images from the COCO dataset of val2017 set as shown in Fig. \ref{quality}. The results indicate that our method is applicable in in-the-wild cases.

\subsection{Running Time Comparisons with SOTA}
Table \ref{tab:Run-time} reports the Running Time Comparisons of our approach and the existing
state-of-the-art approaches. In this experiment, the running time is measured on a single RTX 2070 Super GPU, Intel i7-10700 CPU, and 32 GB RAM with batch size 1.
Since the multi-task structure of \cite{zhen2020smap} and our method, the run times of the bottom show great superiority over any top-down method. As shown in Tab. \ref{tab:Run-time}, our method achieves superior performance, significantly faster than the competing methods. Additionally, our method processing time is roughly constant regardless of the number of people and is more efficient in terms of model size (\#Params) and computation complexity (GFLOPs) as shown in Fig \ref{tab:Run-time}.

\begin{table}
\centering
\begin{spacing}{1.15}
\begin{tabular}{llll}
\hline
\makecell[c]{Method}  &  \makecell[c{p{0.9cm}}]{$\text{MRPE}_x$} & \makecell[c{p{0.9cm}}]{$\text{MRPE}_y$}&\makecell[c{p{0.9cm}}]{$\text{MRPE}_z$}\\
\hline
\makecell[c]{Baseline}            & \makecell[c]{27.5} & \makecell[c]{28.3} & \makecell[c]{261.9}    \\
\makecell[c]{Baseline w/o limb joints} &\makecell[c]{ 24.5 }&\makecell[c]{ 24.9} &\makecell[c]{ 220.2 }     \\
\makecell[c]{Baseline with RANSAC }  &\makecell[c]{ 24.3} & \makecell[c]{24.3} &\makecell[c]{ 207.1 }    \\
\makecell[c]{Moon et al. \cite{moon2019camera}$\ast$}   &\makecell[c]{ 23.3} &\makecell[c]{ 23.0} &\makecell[c]{ 108.1}     \\
\makecell[c]{Lin et al. \cite{lin2020hdnet}$\ast$}    &\makecell[c]{ \bf{15.6}} &\makecell[c]{ \bf{13.6}} & \makecell[c]{69.9}     \\
\makecell[c]{Ours}                 & \makecell[c]{18.8}  &\makecell[c]{22.9 }  & \makecell[c]{ \bf{68.8}}    \\ 
\hline
\end{tabular}
\caption{MRPE results on the single person Human3.6M dataset. Best in \bf{bold}. $\ast$ means using Top-down manner with the help of large detection datasets.}
\label{tab:Human3.6M}
\end{spacing}
\vspace{-3em} 
\end{table}

\subsection{Ablation Studies}
Ablation studies are performed to validate the effectiveness of each branch. To show how the performance of each branch affects the accuracy of the 2.5D pose and absolute depth, we compare the performance of $\text{PCK}_\text{rel}, \text{PCK}_\text{abs}$ and $\text{PCK}_\text{root}$ on each branch separation.

\noindent\textbf{Effect of structured pose estimation.}
As appearance-based heatmap representations have a more precise spatial positioning accuracy, from Table \ref{tab:Ablationstudy}, we see obvious improvement on $\text{PCK}_\text{rel}$ by 2.9\% when only using heatmap representations. Although 2.5D joint offset scheme estimates show compromising ability at positioning accuracy, they show more robustness in inferring occluded joints. The result also shows that the offset scheme can gain an improvement by 1.8\% $\text{PCK}_\text{rel}$. It is able to infer invisible joints and encode the information in a more robust way. The above results show that the performance of our model will degrade severely without our structured Pose Estimation scheme.

\begin{table*}[htbp]
\centering
\begin{spacing}{1.15}
\begin{tabular}{l|lllll}
\hline
\makecell[c]{Method}  & 
\makecell[c]{Input size} & \makecell[c]{$\text{Times(ms)-3 people}$} &  \makecell[c]{$\text{Times(ms)-20 people}$} & \makecell[c]{$\text{FLOPs(G)}$}& \makecell[c]{$\text{Params(M)}$}\\
\hline
\multicolumn{6}{c}{\emph{Top-down methods}}\\
\makecell[c]{Moon et al. \cite{moon2019camera}}         &\makecell[c]{512 $\times$ 512}   & \makecell[c]{169.4}& \makecell[c]{663.1} & \makecell[c]{331.19} & \makecell[c]{112.51}  \\
\makecell[c]{Lin et al. \cite{lin2020hdnet}} &\makecell[c]{512 $\times$ 512}&\makecell[c]{132.9}& \makecell[c]{452.2} &\makecell[c]{306.73} &\makecell[c]{105.91}    \\
\hline
\multicolumn{6}{c}{\emph{Bottom-up methods}}\\
\makecell[c]{Zhen et al. \cite{zhen2020smap}}  &\makecell[c]{832 $\times$ 512}&\makecell[c]{95.6}& \makecell[c]{96.8} & \makecell[c]{176.98} &\makecell[c]{91.48}   \\
\makecell[c]{\bf{Ours}}                 &\makecell[c]{832 $\times$ 512} &\makecell[c]{\bf{43.2}} & \makecell[c]{\bf{43.5}} &\makecell[c]{\bf{112.87}}  & \makecell[c]{\bf{32.04}}  \\ 
\hline
\end{tabular}
\caption{Runtime comparisons on a 2070S GPU. Best in \bf{bold}.}
\label{tab:Run-time}
\end{spacing}
\vspace{-2em} 
\end{table*}

\noindent\textbf{Effect of geometry-based backward.} By building a loop from 2.5D pose to depth, we have deduced a geometry-aware depth from 2.5D pose, which can help the network to the better penalty the 2.5D representation learning. Notably, the geometry depth information is used to further improve the representation learning of 2.5D pose estimation in an end-to-end differentiable manner. Table \ref{tab:Ablationstudy} shows that geometry-based backward pass is able to improve both relative and absolute pose estimation by 4.1\% $\text{PCK}_\text{root}$ and 3.6\% $\text{PCK}_\text{root}$ improvement. 

\noindent\textbf{Effect of adaptive fusion.} In Section \ref{sec:Differentiable}, we developed a depth fusion scheme that leverages both direct visual perception and geometry to alleviate inherent depth ambiguity issues. $\text{PCK}_\text{root}$ is the most direct metric to reflect the ability of our model for absolute depth localization. As shown in Table \ref{tab:Ablationstudy}, our scheme shows significant improvement on $\text{PCK}_\text{root}$ by $5.8\%$. Since the training and test images of the MuPoTS-3D dataset are respectively from synthesized and outdoor, the improvement is more significant in this dataset, which further validates the robustness for generalization of our method.

\section{Conclusion}
In this paper, we developed a unified bottom-up model that leverages the mutual benefits of both 2.5D pose and depth estimation to handle the Monocular 3D-MPE problem. Different from existing top-down or bottom-up methods that treat camera-centric 3D-MPE as two unrelated subtasks: 2.5D pose representation and absolute depth estimation, our method can bridge the gap between 2.5D pose representation and depth estimation and thus benefit from each other. First, we designed a robust structured 2.5D pose estimation is designed to recognize inter-person occlusion based on depth relationships. Additionally, we developed an end-to-end differentiable geometry depth reasoning method that exploits the mutual benefits of both 2.5D pose and camera-centric root depths. This method first uses 2.5D pose and geometry information to infer camera-centric root depths in a forward pass, and then exploits the root depths to further improve representation learning of 2.5D pose estimation in a backward pass. Further, we designed a depth fusion scheme that leverages both visual perception and body geometry to alleviate inherent depth ambiguity issues. Extensive experiments demonstrate the superiority of our proposed model over a wide range of bottom-up methods. Our accuracy is even competitive with top-down counterparts. Notably, our model runs much faster than existing bottom-up and top-down methods. 

\begin{table}
\centering
\begin{spacing}{1.15}
\begin{tabular}{llll}
\hline
\makecell[c]{Method}  & \makecell[c]{PCK\_\text{rel}} & \makecell[c]{PCK\_\text{abs}} & \makecell[c]{PCK\_\text{root}} \\
\hline
\makecell[c]{Full model }                     & \makecell[c]{\bf{81.5}}  & \makecell[c]{\bf{39.5}} & \makecell[c]{\bf{48.9}}   \\
\makecell[c]{w/o offset }               & \makecell[c]{79.7}  &\makecell[c]{ 38.2} & \makecell[c]{48.7}   \\
\makecell[c]{w/o heatmap  }                   & \makecell[c]{78.6}  &\makecell[c]{ 38.0} &\makecell[c]{ 48.9}     \\
\makecell[c]{w/o GB backward}     & \makecell[c]{77.4}  &\makecell[c]{ 35.9} & \makecell[c]{44.7}    \\
\makecell[c]{w/o adaptive fusion }            & \makecell[c]{81.4}  & \makecell[c]{35.1} &\makecell[c]{ 43.1}   \\
\hline
\end{tabular}
\caption{Ablation study of the structure design on the MuPoTS-3D dataset. GB backward stand for geometry-based backward. Best in \bf{bold}.}
\label{tab:Ablationstudy}
\end{spacing}
\vspace{-3em}
\end{table}


\section{Acknowledgements}
This work was supported by the Shanghai Sailing Program (21YF1429400, 22YF1428800), Shanghai Local college capacity building program (22010502800), NSFC programs (61976138, 61977047), the National Key Research and Development Program (2018YFB2100500), STCSM (2015F0203-000-06), and SHMEC (2019-01-07-00-01-E00003).

\bibliographystyle{ACM-Reference-Format}
\bibliography{ACMMM2022-arXiv}

\appendix

\end{document}